\documentclass[letterpaper, 10 pt, journal]{IEEEtran}
\IEEEoverridecommandlockouts                          
\usepackage{graphics} 
\usepackage{epsfig}
\usepackage{times}
\usepackage{amsmath}
\usepackage{amssymb}
\usepackage{threeparttable}
\usepackage{booktabs}
\usepackage{soul}
\usepackage{subfigure}
\usepackage{xcolor}
\usepackage[hidelinks]{hyperref}
\usepackage{algorithm}
\usepackage{algorithmic,comment}
\usepackage{tcolorbox}
\usepackage{multirow}

\title{\LARGE \bf
Vision-Language-Action Models for Selective Robotic Disassembly: \\A Case Study on Critical Component Extraction from Desktops
}

\author{Chang Liu$^{1,*}$, Sibo Tian$^{1,*}$, Sara Behdad$^{2}$, Xiao Liang$^{3,\dagger}$, and Minghui Zheng$^{1,\dagger}$
\thanks{This work was supported by the USA National Science Foundation under Grant No. 2422826 and No. 2026276. Portions of this research were conducted with the advanced computing resources provided by Texas A\&M High Performance Research Computing.}
\thanks{$^{1}$ Chang Liu, Sibo Tian and Minghui Zheng are with the J. Mike Walker '66 Department of Mechanical Engineering, Texas A\&M University, College Station, TX 77843, USA. {\tt\small Emails: {changliu.chris, sibotian, mhzheng}@tamu.edu.}}
\thanks{$^{2}$ Sara Behdad is with the Engineering School of Sustainable
Infrastructure \& Environment, University of Florida, Gainesville, Florida 32611, USA. {\tt\small Email: sara.behdad@essie.ufl.edu.}}
\thanks{$^{3}$ Xiao Liang is with the Zachry Department of Civil and Environmental Engineering, Texas A\&M University, College Station, TX 77843, USA. {\tt\small Email: xliang@tamu.edu.}}
\thanks{$^*$ Equal Contribution.}
\thanks{$^\dagger$ Corresponding Authors.}}

\begin{document}

\maketitle
\thispagestyle{empty}
\pagestyle{empty}

\begin{abstract}

Automating the selective disassembly of critical components from end-of-life (EoL) desktops, such as high-value items like RAM modules and CPUs, as well as sensitive parts like hard disk drives, remains challenging due to the inherent variability and uncertainty of these products. Moreover, their disassembly requires sequential, precise, and dexterous operations, further increasing the complexity of automation. Current robotic disassembly processes are typically divided into several stages: perception, sequence planning, task planning, motion planning, and manipulation. Each stage requires explicit modeling, which limits generalization to unfamiliar scenarios and may lead to cumulative errors. The recent development of vision-language-action (VLA) models has presented an end-to-end approach for general robotic manipulation tasks. The direct translation of high-level visual perception and natural language instructions into robot motion and manipulation may minimize or even eliminate the need for intermediate stages. Although VLAs have demonstrated promising performance on simple tasks, the feasibility of applying such models to the complex disassembly domain remains largely unexplored. In this paper, we collected a customized UR5e demonstration dataset specifically for RAM and CPU disassembly and used it to fine-tune two well-established VLA approaches, OpenVLA and OpenVLA-OFT, as a case study. We divided the whole disassembly task into several small steps, and our preliminary experimental results indicate that the fine-tuned VLA models can faithfully complete multiple early steps but struggle with certain critical subtasks, leading to task failure. However, we observed that a simple hybrid strategy that combines VLA with a rule-based controller can successfully perform the entire disassembly operation. These findings highlight the current limitations of VLA models in handling the dexterity and precision required for robotic EoL product disassembly. By offering a detailed analysis of the observed results, this study provides insights that may inform future research to address current challenges and advance end-to-end robotic automated disassembly.

\end{abstract}

\section{Introduction}

\begin{figure}[t]
    \begin{center}
        \includegraphics[width=0.45\textwidth]{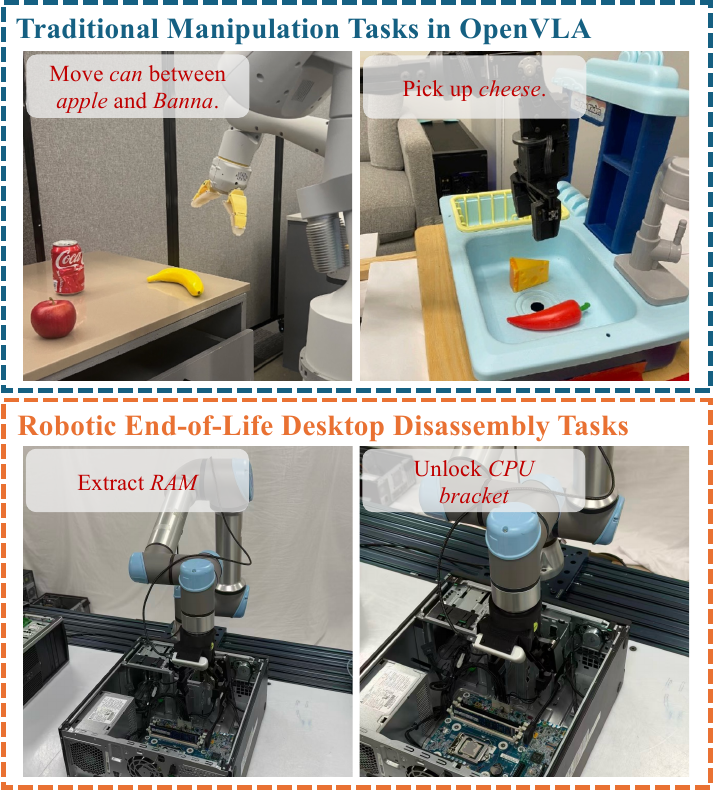}
    \end{center}
        \caption{Manipulation task comparison. In most traditional VLA applications, the environment is clean and structured, and the target object is visually distinct. However, in robotic disassembly, the target is often small, difficult to isolate visually, and demands a much higher level of precision. }
    \label{figure:comparison} 
    \vspace{-0.1in}
\end{figure}

Electronic waste (e-waste) is one of the fastest-growing waste streams in the world \cite{shahabuddin2023review}, driven by rapid technological advances and frequent product upgrades. Global e-waste generation was recorded at 62 million tonnes in 2022 and is expected to escalate to 82 million tonnes by 2030 \cite{balde2024global}. However, less than 25\% of e-waste is properly processed, resulting in massive amounts of raw materials and high-value components being directly discarded in landfills \cite{jain2023review}, despite e-waste being both environmentally hazardous and a valuable source of recoverable materials \cite{liu2023global}. Properly recycling end-of-life (EoL) electronics has therefore become an increasingly urgent global challenge due to the growing environmental and economic concerns. 

Disassembly plays a fundamental role in the e-waste recycling process, enabling the extraction of valuable components, enhancing recovery rates, and ensuring environmental safety through the proper handling of hazardous materials. However, modern electronic products, such as desktops, exhibit high levels of customization, which significantly complicates the disassembly process. Additionally, the inherent variability of EoL desktops, in terms of both type and usage conditions, requires sophisticated dynamic decision-making and planning skills throughout the entire disassembly process. Consequently, manual disassembly remains the primary solution in practice, as humans can handle the associated uncertainty during the disassembly process. However, it is extremely labor-intensive and time-consuming, making it impractical to align with the rapidly growing volume of EoL electronic products.

With recent advancements in machine learning and robotics, recycling practices have increasingly incorporated intelligent robots to enhance automation levels, which significantly alleviates the stress of the growing labor shortage and improves the efficiency of e-waste recycling. Moreover, research studies have proposed human-robot collaborative disassembly (HRCD) as a promising approach \cite{lee2024review}, combining human adaptability with robotic efficiency to address the complexities of EoL product recycling. While HRCD offers an effective intermediate solution to the current recycling practice, achieving fully automated disassembly for EoL products without a human in the loop still remains a long-term goal \cite{hellmuth2021assessment, liu2026raise}. However, the capabilities of robots throughout the entire process present a key bottleneck in current disassembly practices. The challenge lies not only in how robots can handle the uncertainty of EoL products, but also in determining and executing the appropriate actions once the relevant information has been processed.

Current research on robotic disassembly is typically divided into different stages. These include utilizing computer vision and perception systems to recognize and localize the components, disassembly sequence planning to define the order of the tasks, and task planning to distribute the work to different agents regarding their capabilities. Once the robot receives the task, it also needs to perform motion planning and manipulation to complete the task. Each of these stages requires a specialized model tailored to its specific function, which may not be optimized to integrate seamlessly together and may lead to cumulative errors. 

Vision-language-action (VLA) models, which map visual inputs and task descriptions directly into robotic actions through an end-to-end framework \cite{kawaharazuka2025vision}, offer an impressive alternative to traditional stage-separated learning-based solutions. These models have been explored primarily in simple daily-life tasks, demonstrating strong adaptive capabilities for robot agents. However, the feasibility of applying VLA models to more complex scenarios, such as e-waste disassembly, where the operational dexterity and precision are critical and visual inputs are highly cluttered with overlapping and disorganized electronic components, has yet to be investigated. The difference of these tasks can be seen in Figure~\ref{figure:comparison}.

\begin{figure*}[htbp]
    \begin{center}
        \includegraphics[width=1.0\textwidth]{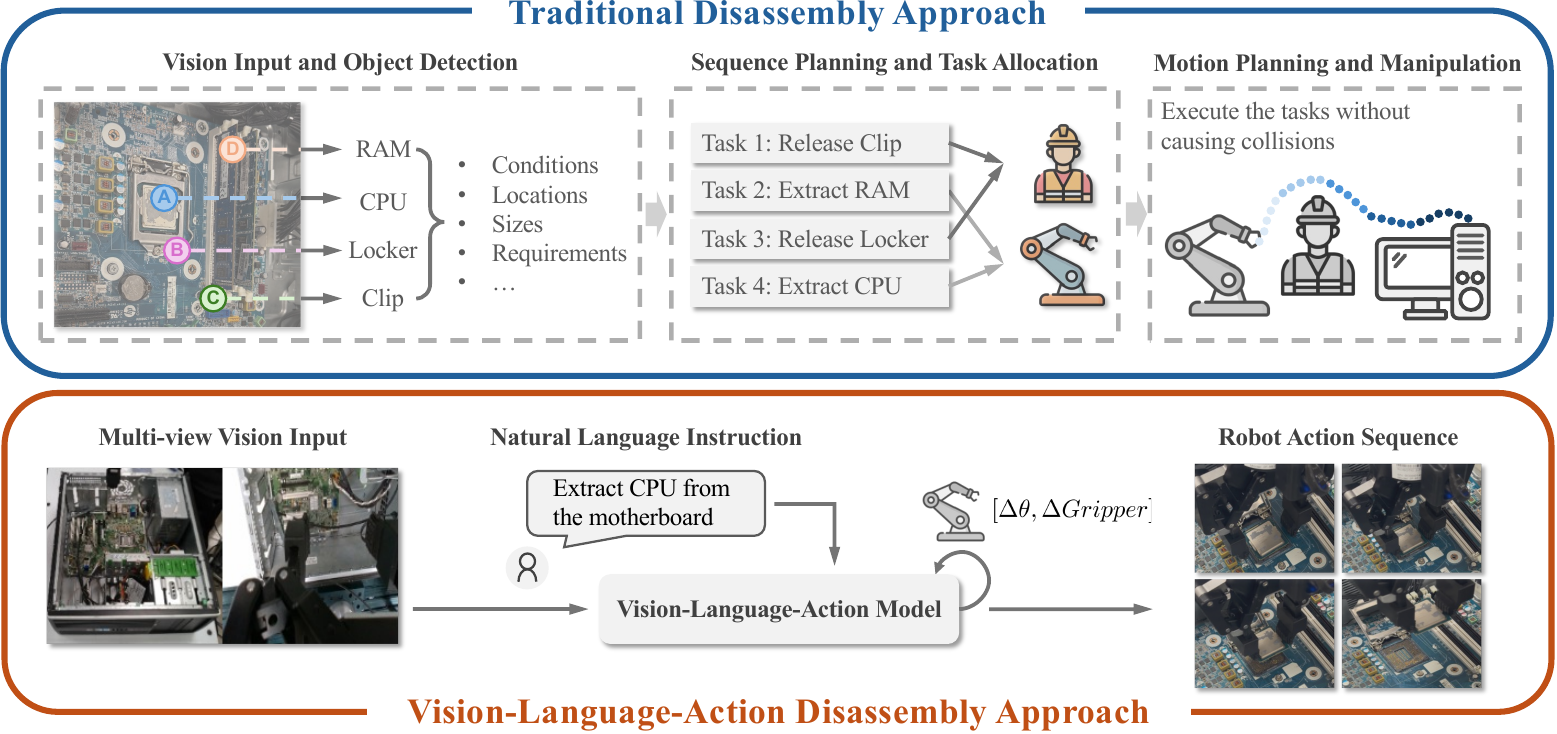}
        \caption{Comparison between traditional multi-stage disassembly approaches and end-to-end vision–language–action methods.}
    \label{figure_end2end} 
    \end{center}
    \vspace{-0.1in}
\end{figure*}

In this paper, we present a case study to evaluate the feasibility and potential of utilizing VLA models in complex robotic disassembly tasks. We focus on selective disassembly of high-value components by providing different natural language instructions, leveraging vision systems, and generating corresponding actions for robots to execute. To this end, we first built a teleoperation platform for data collection and then conducted two disassembly operations: RAM removal and CPU bracket unlocking. Based on these teleoperated demonstrations, we constructed a customized robotic vision-language-action dataset to fine-tune two well-established VLA models, OpenVLA \cite{kim2024openvla} and OpenVLA-OFT \cite{kim2025fine}, and assess their performance in complex disassembly tasks. Although neither model can complete the entire disassembly task, both were able to guide the robot through several early steps of the task. By combining VLA with a simple rule-based control strategy, the framework can successfully execute the full disassembly. These results demonstrate that VLA models can offer valuable high-level guidance in disassembly, with failures occurring primarily in stages that require precise localization and contact. This highlights the current limitations of VLA approaches when applied to contact-rich and dexterous disassembly tasks. We further provided a detailed analysis to offer insights for future research on VLA applications in end-to-end robotic disassembly systems.

\section{Literature Review}

Complete disassembly involves taking apart all components of an EoL product. However, it is time-consuming and usually unnecessary. In contrast, selective disassembly focuses only on components with high recycling value, striking a balance between efficiency and economic benefit. As a result, selective disassembly has become a practical and effective approach in modern semi-automated and automated disassembly systems. Recent research has examined multiple stages of the disassembly workflow to identify pathways that better support robotic participation in disassembly operations \cite{lee2024review}.

The perception module typically incorporates deep learning-based computer vision techniques to improve the localization of critical components within EoL products \cite{zhou2024towards}. Jahanian et al. \cite{Jahanian_2019_CVPR_Workshops} proposed an instance segmentation model to detect the boundaries of small electronic components on phone PCB boards. Zhang et al. \cite{zhang2023automatic} presented a two-stage model that locates and extracts screws from image input. However, these learning-based methods focus on specific tasks and are limited by the availability of training data. These constraints affect the performance and limit the applicability of such models in general practice \cite{dong2025benchmarking}. To address these limitations, Deng et al. \cite{deng2025learning} proposed a model that learns feature representations from new products and generalizes to detect various damaged components in EoL products.

The results from the perception system can be further leveraged to guide sequence, task, and motion planning in robotic disassembly. The current disassembly sequence and task planning focus on improving overall efficiency by allocating tasks between human and robot operators based on their capabilities in the HRCD scenarios \cite{lee2024review, he2024disassembly}. Lee et al. \cite{lee2022task} proposed a task allocation and sequence planning algorithm for hard disk drive disassembly that optimizes efficiency while ensuring safety constraints. Peng et al. \cite{peng2025dynamic} introduced an agent-based deep reinforcement learning approach for sequence planning to address human behavioral uncertainty in practice. Guo et al. \cite{guo2023human} presented an optimized HRCD sequence that considers failure characteristics, thereby improving the overall disassembly efficiency. However, these algorithms do not fully utilize the detection results from the vision system for real-time planning. Most algorithms predefine the product's conditions and structure, leaving a significant gap between current research study and practical requirements. A recent study by Yu et al. \cite{yu2025rescheduling} utilized a multi-modal large language model with knowledge graphs to perform sequence planning and task allocation directly from real-time vision feedback. This approach offers a novel way to enhance both generalization capability and real-time planning reliability in dynamic environments.

Once a task is allocated to a robot agent, it needs to plan a trajectory to operate safely for robotic disassembly \cite{asif2024robotic}. Motion planning has been extensively explored with particular emphasis on safety and efficiency \cite{liu2024hybrid, tian2025warm, soleymanzadeh2025simpnet}. While these works demonstrate strong performance, they typically assume static environments, whereas HRCD scenarios need to account for human agents moving within the shared workspace. To address this limitation, researchers have proposed leveraging human action recognition \cite{zhang2024early} and motion prediction algorithms \cite{tian2024transfusion, zhang2025multi, tian2025real} to anticipate human behavior. The forecasted human motion can be integrated into the planning process \cite{liu2023task}, enabling the robot to proactively plan its motion. Additionally, Song et al. \cite{song2025adaptive} proposed an adaptive motion planning framework to ensure the safety of human operators through physical interaction while working with a robot. 

After the robot executes the planned trajectory and moves to the target region, a key challenge is enabling the robot to autonomously perform the required manipulation to complete the disassembly task without human intervention, which is an essential capability for advancing robotic disassembly systems. Manipulation tasks have been investigated broadly through imitation learning \cite{fang2019survey} and reinforcement learning (RL) \cite{elguea2023review}. Ankile et al. \cite{ankile2024juicer} proposed an imitation learning framework with limited human demonstrations for assembly tasks. However, the robustness of imitation learning under environmental variability remains limited, particularly in disassembly scenarios \cite{foo2022artificial}. Luo et al. \cite{luo2025precise} introduced a vision-based human-in-the-loop RL framework, enabling robots to learn complex manipulation skills from human demonstrations within a practical training time. Zang et al. \cite{zang2025robotic} presented that external knowledge injected into RL can enhance learning performance for robotic disassembly tasks. Although RL generally demonstrates better performance over imitation learning, it remains highly task-specific and requires retraining for each new task, which restricts the generalization of robotic manipulation in real-world applications. 

Recent progress in VLA models provides a promising direction toward end-to-end robotic manipulation by integrating visual perception and semantic understanding through large-scale pretrained models, leading to stronger generalization across different environments and task settings. The difference of these two methods can be seen in Figure~\ref{figure_end2end}. Several VLA models have garnered broader attention, including RT2 \cite{zitkovich2023rt}, OpenVLA \cite{kim2024openvla}, and $\pi_{0.5}$ \cite{intelligence2025pi_}, which have demonstrated impressive generalization capabilities in real-world robotic tasks. However, their use in industrial-level operations, such as disassembly, which involve constrained geometry, contact-rich manipulation, and tight precision requirements, remains largely unexplored.

\section{Experimental Studies and Results}

In this work, we fine-tuned two widely used frameworks, OpenVLA \cite{kim2024openvla} and OpenVLA-OFT \cite{kim2025fine}, to evaluate the potential and limitations of VLAs in a complex EoL product disassembly task. While both models have demonstrated promising performance in tabletop manipulation and everyday object-handling scenarios, their performance in long-horizon, contact-rich industrial operations remains largely unexplored. Therefore, we considered two robotic selective disassembly tasks for EoL desktops in this study: RAM module removal and CPU bracket unlocking. 

RAM module removal can be viewed as a pickup operation; however, precise grasping is critical in this case, as the RAM is tightly installed on the motherboard and the two-finger gripper must move to a very specific location to securely grasp the thin module without disturbing nearby components. On the other hand, unlocking the CPU bracket represents the most contact-rich and dexterous operation in desktop disassembly. It requires the robot to precisely manipulate a small metal lever, apply force in the correct direction, and follow a constrained motion sequence to disengage the latch without damaging the surrounding socket or motherboard. The fine-grained motion required by these two disassembly tasks makes them significantly more challenging than typical tabletop manipulation scenarios.

\begin{figure}[t]
    \begin{center}
        \includegraphics[width=0.5\textwidth]{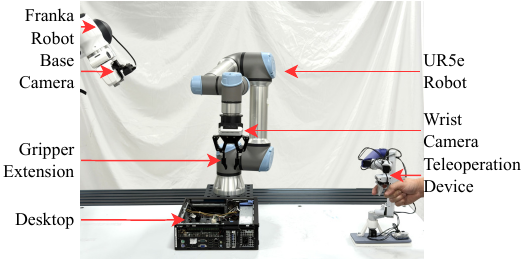}
    \end{center}
        \caption{The teleoperation setup for imitation data collection.}
    \label{figure:Setup} 
\end{figure}

\subsection{Data Collection}

\begin{figure*}[t]
    \begin{center}
        \includegraphics[width=1.0\textwidth]{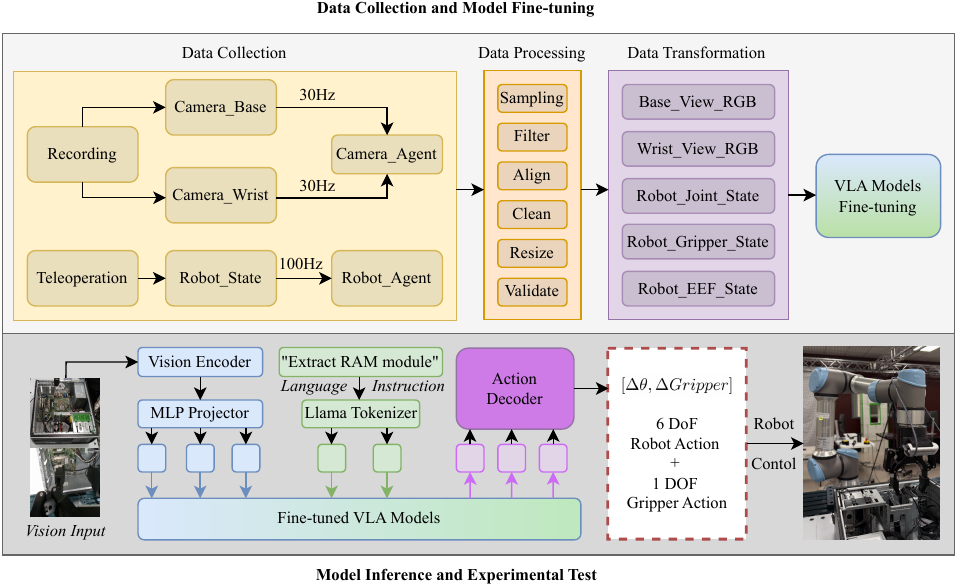}
    \end{center}
        \caption{The data processing steps for fine-tuning models and the utilization of VLA models in experimental tasks.}
    \label{figure: Data} 
\end{figure*}

To collect high-quality disassembly demonstrations, we built a teleoperation platform based on the Gello system \cite{wu2024gello}. The original design was adapted to integrate with the UR5e robot and to ensure the structural robustness during operation. Additionally, we designed an extension for the Robotiq 2F-85 gripper, featuring a specialized column that helps stabilize the bracket lever during unlocking. This design improves manipulation stability in desktop disassembly tasks. Unlike DeGrip \cite{zhang2025degrip}, which is a gripper explicitly designed for these tasks, our extension works with standard two-finger grippers and enhances their applicability in complex disassembly tasks.

For vision recording, two RGB-D cameras were used to capture complementary views of the task. An OAK-D camera was mounted on a Franka Panda end effector to provide a stable base view, serving as a replicable reference for testing. A RealSense D435i was fixed to a custom mount designed for the UR5e equipped with the 2F-85 gripper. The mount securely holds the camera at a $30^\circ$ angle, ensuring the tip of the gripper extension remains visible throughout the operation. The whole setup is shown in Figure~\ref{figure:Setup}. 

We recorded robot states and images asynchronously to avoid blocking during data collection and later aligned them using timestamps. Both cameras operated at 30 Hz with low-resolution output to conserve storage, and each RGB frame was stored together with corresponding depth information. To guarantee the smoothness of robot motion, the communication frequency between the controller and teleoperation system was set to 100 Hz. However, due to the hardware communication constraints, the effective control rate was approximately 48 Hz. All robot states, including joint angles, end effector positions, and gripper states, were stored accordingly. We used 10 different desktops with varying conditions and configurations for disassembly data collection. For the RAM module removal task, 164 demonstrations were recorded, with 158 successfully completed on the first attempt. The CPU bracket unlocking task included 123 demonstrations, of which 110 were completed on the first attempt. A total of 287 trajectories were recorded across two disassembly tasks. 

To process the recorded asynchronous data, robot states were first downsampled to match the timestamps of the base-camera frames. We removed redundant robot state entries and aligned the wrist-camera images with the base-view images. All files associated with the same demonstration were merged into one pickle file. RGB images from two cameras were resized to a resolution of 224 x 224 to meet the fine-tuning requirements of the selected pretrained VLA models. A final check was performed to verify the file structure before data transformation. The whole process is shown in Figure~\ref{figure: Data}.

\begin{figure*}[t]
    \begin{center}
        \includegraphics[width=1.0\textwidth]{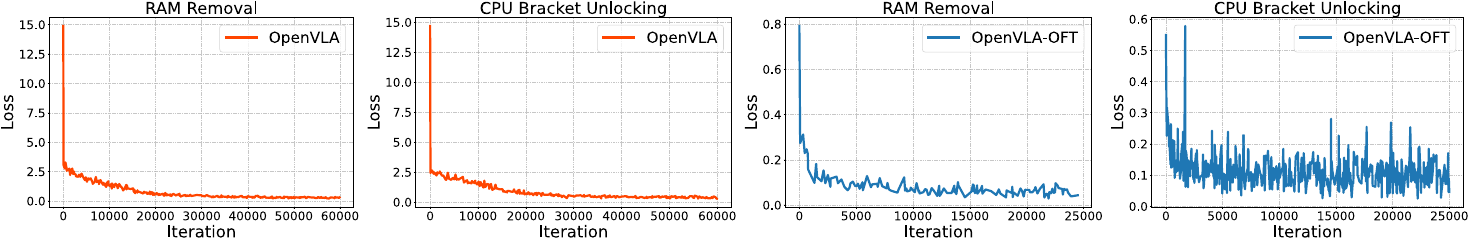}
    \end{center}
        \caption{Training loss of two VLA models on two disassembly tasks.}
    \label{loss} 
    \vspace{-0.1in}
\end{figure*}

\subsection{Fine-tuning}

We selected two representative VLA models and fine-tuned them on our disassembly dataset in this work. OpenVLA \cite{kim2024openvla} is a 7B-parameter manipulation policy built on Llama2 \cite{touvron2023llama}. It discretizes continuous robot motions into discrete robot action tokens and overwrites the least frequent tokens in the Llama2's original codebook to accommodate these action tokens. During training, OpenVLA employs next-token prediction with cross-entropy loss. For inference, it takes a single fixed-view image of the current step and a language instruction as input and autoregressively predicts seven discrete robot action tokens, six for each joint and one for gripper control, for the next timestep.  OpenVLA-OFT \cite{kim2025fine} improves upon OpenVLA by incorporating additional image inputs, parallel decoding, continuous action representation, and action chunking. Specifically, regarding input perspectives, OpenVLA-OFT provides the language decoder with an additional wrist-view image alongside the third-person view camera image, as well as the robot’s proprioceptive states. To obtain continuous actions, OpenVLA-OFT passes the final hidden states to a separate action head and directly maps the hidden states to normalized continuous actions. Consequently, OpenVLA-OFT can directly supervise L1 regression from continuous actions. Additionally, OpenVLA-OFT predicts and executes a sequence of future actions, while OpenVLA can only generate an action for one single timestep in an autoregressive fashion. We leveraged the LoRA fine-tuning \cite{hulora} to efficiently adapt the pretrained VLA models to our disassembly tasks. We trained both models on both tasks, and all models converged to a small loss value, as shown in Figure~\ref{loss}.

\begin{figure*}[t]
    \begin{center}
        \includegraphics[width=0.95\textwidth]{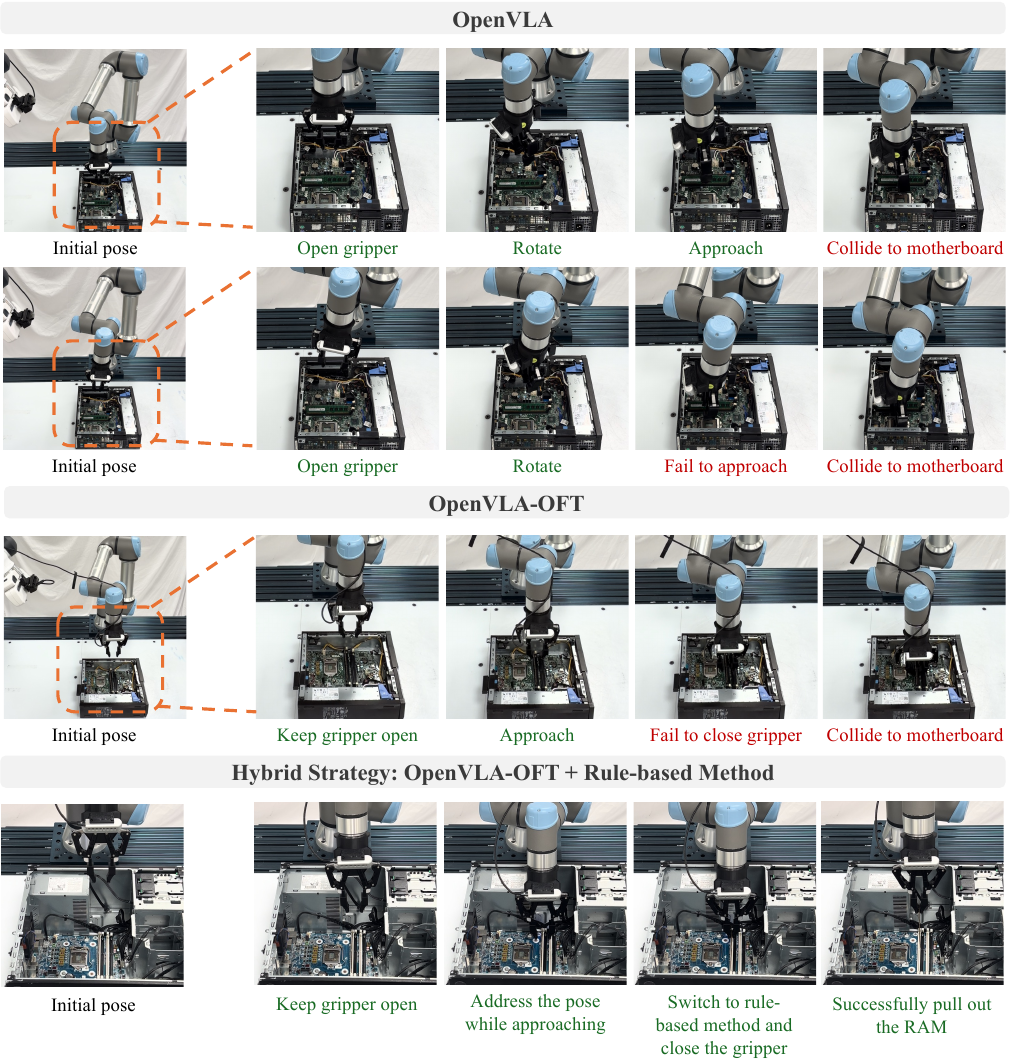}
    \end{center}
        \caption{Experimental results for different VLA models and control strategies.}
    \label{demo} 
    \vspace{-0.1in}
\end{figure*}

\subsection{Tests and Results}

\begin{table*}[htbp]
    \caption{Performance of fine-tuned VLA models across different stages of two selected disassembly tasks.}
    \centering
    \resizebox{1.0\textwidth}{!}{
    \begin{threeparttable}
    \begin{tabular}{cccccc|cccc}
        \toprule
        \multirow{2}{*}{Model} 
        & \multicolumn{5}{c|}{Task 1: RAM Removal}
        & \multicolumn{4}{c}{Task 2: CPU Unlocking} \\ \cmidrule{2-10}
        & Alignment & Approach & Localization & Configuration & Actuation 
        & Approach & Localization & Configuration & Actuation \\ \midrule
        OpenVLA & 12/20 & 13/20 & 4/20 & 4/20 & 0/20 & 1/20 & 0/20 & 12/20 & 0/20 \\ 
        \midrule
        OpenVLA-OFT & 16/20 & 16/20 & 7/20 & 19/20 & 0/20 & 20/20 & 0/20 & 20/20 & 0/20 \\ \bottomrule
    \end{tabular}
    \begin{tablenotes}
        \item[*] \textit{Alignment} refers to the robot's ability to align the gripper orientation with the target module. \textit{Approach} measures whether the robot can move towards the target module. \textit{Localization} means whether the robot can reach a position that is precise enough for actuation. \textit{Configuration} checks whether the gripper can maintain the correct state that is prepared for actuation. \textit{Actuation} is the final manipulation motion for the target module disassembly.
    \end{tablenotes}
    \end{threeparttable}
    }
    \label{tab:vla_disassembly_breakdown}
\end{table*}

We evaluated two fine-tuned VLA models on the proposed disassembly tasks using an evaluation setup that replicated the data-collection environment, allowing the models to operate under camera configurations and operating conditions matched as closely as possible. These disassembly tasks introduce complex object layout and spatial constraints that rarely appear in other tabletop manipulation benchmarks. The evaluation aimed to see how the trained policies behave when transferred to real-world disassembly, where visibility and precision constraints become the dominant bottlenecks.

Empirically, we found that utilizing the fine-tuned VLA models alone could not complete any trials on either disassembly task. For OpenVLA, this training-to-test performance drop is expected under the single-camera setting, in addition to the data covariate shift. The policy relies entirely on the base-camera view, where the target is frequently occluded as the end effector moves into the desktop during the final approaching stage. Once the target drops out of view, the model cannot reliably predict the next action for controlling the robot, making single-view perception difficult to sustain through the final approach. Despite this, the robot's behavior indicates that OpenVLA partially learned the skills required for disassembly tasks. In the RAM extraction trials, the robot could approach the RAM module with some orientation adjustment, then it entered a repeated up–down pattern near the slot and tried to grasp the RAM. This motion behavior suggests that the robot attempted to interact with the RAM. However, it lacked a sense of whether the gripper had securely held the RAM module and whether the pull-out action had finished, as the robot's action prediction depended solely on a single side-camera view. After several such attempts, the robot caused some unexpected collisions with the motherboard, and none of the trials resulted in a successful RAM extraction. The CPU disassembly task showed a similar level of partial competence. The robot closed the gripper and moved downward correctly at the beginning. However, it could not precisely find the narrow contact point, which is required to operate the lever locker. In summary, these patterns indicate that the model captured the general structure of the tasks but struggled to maintain consistent and reliable behavior. Besides the test-time covariate shift, this difficulty may also result from information loss caused by the single-camera setup and the lack of robot proprioceptive state input. Such limitations also make it essential to examine whether the second view, robot proprioceptive input, and multi-step prediction can improve the overall stability.

As for the OpenVLA-OFT model, the inclusion of an additional wrist-view camera, robot proprioceptive state input, and action chunking led to more pronounced improvements in the RAM task. The robot was able to track the orientation of the module more precisely and adjust the gripper pose when the relative angle changed. It could also approach the slot with less hesitation, suggesting that the second view helped preserve target visibility during the approach. However, none of the trials resulted in a successful removal. The grasp attempts were consistently close but still incomplete, and the final extraction motion was not precise enough to release the module. The improvements were less apparent in the CPU task. Although the robot consistently repeated the correct initial steps, it still failed to locate the narrow contact point required to operate the lever. These additional enhancements were insufficient to meet the precision demands of this task. Overall, the OpenVLA-OFT model demonstrated more coherent behavior than the base OpenVLA model, but its performance remained limited in both tasks. The experimental results are shown in Figure~\ref{demo}.

To provide a more structured view of intermediate behaviors, we further evaluated both models across the key stages of two tasks rather than focusing solely on the final success rate. This stage-wise analysis reveals which steps in the disassembly process the VLA models can reliably complete and where their performance begins to degrade. These insights highlight the parts of the tasks that remain particularly challenging for current VLA approaches and raise the question of whether a simple low-level control strategy might be more effective in those stages. Specifically, for the RAM removal task, we evaluated the execution across five subtasks. Alignment refers to the robot's ability to align the gripper orientation with the RAM module. Approach measures whether the robot can move towards the RAM module. Localization means whether the robot can reach a position that is precise enough for actuation. Configuration checks whether the gripper can maintain the correct state that is prepared for actuation. Actuation is the final extraction motion for the RAM module. We conducted 20 experiments with each model, and presented the results in Table~\ref{tab:vla_disassembly_breakdown}. The OpenVLA model aligned with the RAM module in 12 trials and approached it correctly in 13 trials. It precisely localized the RAM in only 4 trials. The gripper also reached the desired configuration for narrow space extraction in 4 trials. The OpenVLA-OFT model performed better across the same steps. It aligned with the RAM module in 16 trials and approached it correctly in 16 trials. It precisely localized the RAM in 7 trials and maintained the desired gripper configuration in 19 trials. However, neither model could achieve the final actuation to extract the RAM module. As for the CPU unlocking experiment, there is no alignment step because the gripper was already configured to maintain the required orientation throughout the task. The OpenVLA model moved toward the CPU lever only once in all 20 trials. It kept the gripper in the required closed configuration for unlocking in 12 trials. The OpenVLA-OFT model demonstrated clear improvement, accurately approaching the lever in all trials while maintaining the correct gripper configuration. However, neither model was able to precisely localize the lever position and further complete the unlocking actuation.

Since neither VLA model could complete the actuation stage, we examined whether introducing a low-level control strategy could enable task completion. We conducted an additional test to see what happens when the VLA models handle only the high-level decisions and motions. Once the robot reached the desired height, a simple position controller executed the gripper configuration and the actuation. This hybrid strategy is motivated by the observation that the robot needs intelligent decision-making skills to precisely locate the component and move toward it, but once it reaches the target pose, the remaining task is simply to close the gripper firmly and lift the component. We ran 10 tests on the RAM extraction task, and 2 of them succeeded. This indicates that direct position control can reliably complete the final actuation motion as long as the robot can precisely reach the required pose. The same method did not help with the CPU task, because the robot never reached the position needed to unlock the lever. These results illustrate that the main bottleneck for the VLA models in the disassembly scenes is precision.

\section{Conclustion}

This study examined the performance of two well-established VLA models after fine-tuning them on our disassembly dataset. Both models were able to reproduce portions of the disassembly process and execute several key motions, indicating that the fine-tuning captured certain aspects of the task. However, the overall behavior during testing revealed a clear gap between the training results and the actual disassembly performance in real-world tests. A significant factor behind this difference is covariate shift caused by limited demonstration coverage. When execution deviates from the demonstrated trajectories, the robot finds it difficult to correct its behavior. This indicates that the dataset was insufficient to support fine-tuning for VLA models. In addition to the data limitation, the models' capabilities were further constrained by the low-resolution visual input and the complexity of the desktop environment, where the target components are often difficult to discern. Such challenges are particularly critical to handle precise and dexterous disassembly tasks.

Future work can focus on improving the reliability and performance of VLA models in disassembly. Larger and more diverse demonstrations are needed to accommodate variations across EoL hardware configurations and to mitigate the covariate shift that we observed during testing. Force or tactile signals can also be included in the robot's state. Such information may help the model stay aware of its progress during the task. Additionally, methods such as reinforcement learning can be integrated into the VLA models to correct the robot's behavior and recover from states that fall outside the demonstration distribution.

\bibliographystyle{IEEEtran}
\bibliography{ref}{}

\end{document}